%% file: acl_latex.tex
\pdfoutput=1

\documentclass[11pt]{article}

\usepackage{acl}

\usepackage{times}
\usepackage{latexsym}
\usepackage{csquotes}
\usepackage{subcaption}
\usepackage{multirow}
\usepackage{pgfplots,pgfplotstable}
\pgfplotsset{compat=1.18}
\tikzset{
  font={\fontsize{8pt}{10}\selectfont}}

\usepackage[T1]{fontenc}

\usepackage[utf8]{inputenc}

\usepackage{microtype}

\title{Direct Models for Simultaneous Translation and Automatic Subtitling: FBK@IWSLT2023}

\author{Sara Papi$^{\Diamond,\Box}$, Marco Gaido$^{\Diamond}$, Matteo Negri$^{\Diamond}$ \\
  $^{\Diamond}$Fondazione Bruno Kessler \\
  $^{\Box}$University of Trento \\
  \texttt{\{spapi,mgaido,negri\}@fbk.eu}}

\begin{document}
\maketitle
\begin{abstract}
This paper describes the FBK's participation in the Simultaneous Translation and Automatic Subtitling tracks of the IWSLT 2023 Evaluation Campaign.
Our submission focused on the use of direct architectures to perform both tasks: for the simultaneous one, we leveraged the knowledge already acquired by offline-trained models and directly applied a policy to obtain the real-time inference; for the subtitling one, we adapted the direct ST model to produce well-formed subtitles and exploited the same architecture to produce timestamps needed for the subtitle synchronization with audiovisual content. 
Our English-German SimulST system shows a reduced computational-aware latency compared to the one achieved by the top-ranked systems in the 2021 and 2022 rounds of the task, with gains of up to 3.5 BLEU. Our automatic subtitling system outperforms the only-existing solution based on a direct system by 3.7 and 1.7 SubER in English-German and English-Spanish respectively.
\end{abstract}

\section{Introduction}

In recent years, 
the advances in natural language processing and machine learning
led to
a surge of interest in developing speech translation (ST) systems that can 
translate
speech from one language 
into text in another language
without human intervention.
Significant progress has been specially made toward
end-to-end ST models \citep{berard_2016,weiss2017sequence} trained to directly translate speech without the intermediate steps of transcription (through automatic speech recognition - ASR) and translation (through machine translation - MT).
Along with this 
growing interest in direct ST, 
also accompanied by a reduction of the performance gap with respect to cascaded architectures \citep{bentivogli-etal-2021-cascade}, 
other trends have emerged thanks to deep learning advancements, which made it possible to deploy direct solutions to perform the task in real-time (i.e. to produce partial translations while continuing to process the input audio) or to automatically generate subtitles for audiovisual content (i.e. pieces of translated text which have to conform to specific spatiotemporal constraints and be synchronized with the video).

The International Workshop on Spoken Language Translation (IWSLT) is playing an important role in advancing the state-of-the-art in these fields by organizing a series of evaluation campaigns \citep{ansari-etal-2020-findings,anastasopoulos-etal-2021-findings,anastasopoulos-etal-2022-findings} focused on simultaneous speech translation (SimulST) and, this year for the first time, automatic subtitling.
These campaigns provide a unique opportunity for researchers to compare their systems against others, share their findings, and 
identify areas for further improvement. 

In this paper, we describe 
FBK's participation in the IWSLT 2023 Evaluation Campaigns \citep{iwslt2023} for simultaneous translation and automatic subtitling. 
Motivated by the promising results reported in previous works \citep{ren-etal-2020-simulspeech,papi2022direct}, our approach is characterized by the use of direct ST models to address both tasks.

For the simultaneous speech-to-text translation (SimulST) task, we participated in the English$\rightarrow$German track and  leveraged an offline-trained direct model without performing any adaptation to the real-time scenario, 
as this has recently been shown not to be necessary to achieve competitive results \citep{papi2022does}. For the automatic subtitling task, we participated in both the English$\rightarrow$German and English$\rightarrow$Spanish tracks by adapting a direct ST model to produce well-formed subtitles and exploiting the same architecture to produce the timestamps needed for their synchronization with audiovisual contents, as in \citep{papi2022direct}.

Our results demonstrate the effectiveness of our approach. In SimulST, the computational-aware latency of our models is lower compared to the winning systems of the last two rounds (2021, and 2022) of the IWSLT SimulST Evaluation Campaign, with gains up to 3.5 BLEU. In automatic subtitling, our systems improve the results reported in \citep{papi2022direct} which, to the best of our knowledge, represents the only-existing solution based on a direct model. Specifically, on average among the various dev sets available for the task, we achieve 3.7 SubER on en-de and 1.7 SubER on en-es.



\section{Applied Direct Models}

For this year's submission, we applied the direct ST models to the two different 
scenarios of simultaneous translation and automatic 
subtitling.

\subsection{Simultaneous Translation}
\label{subsec:simul}

Recent trends in SimulST consist of using offline-trained models for simultaneous inference \citep{papi2022does}. 
There are several motivations for this choice:
\emph{i)} it avoids re-training or building specific architectures for SimulST, saving time and computational resources; \emph{ii)} only one model has to be trained and maintained to perform both 
offline and simultaneous ST;
and \emph{iii)} there is no need to train several 
models, each specialized to support
different latency regimes.

A key aspect of SimulST, also critical when approaching the task with offline models at inference time, is the so-called \textit{decision policy}: the mechanism that
is in charge of deciding whether to read more information or to emit a partial hypothesis.
One of the first and most popular policies
is the wait-k \citep{ma-etal-2019-stacl}, initially introduced for simultaneous MT, and then applied to the speech scenario \citep{ma-etal-2020-simulmt,chen-etal-2021-direct,zeng-etal-2021-realtrans,karakanta-etal-2021-simultaneous}. The wait-k, which prescribes waiting for an initial number of $k$ words before starting to translate, is defined as a \enquote{fixed} policy \citep{zheng-etal-2020-simultaneous} because the decision is taken independently from the source input content. However, 
as the actual information contained in the input  (e.g. in terms of ambiguity, completeness, and syntactic/semantic cohesion) is also important for the sake of good-quality incremental translations,
several \enquote{adaptive} policies have been introduced, which instead adapt their decisions to the input content. Some adaptive policies require 
system re-training or the development of ad-hoc modules
\citep{liu-etal-2021-cross,chang22f_interspeech,zhang2022information}, while some others do not \citep{liu20s_interspeech,nguyen2021empirical,papi2022attention}. Since our objective is to avoid any modifications to the offline-trained model, we pointed our attention to the 
latter, more conservative category. Among these policies, we analyzed 
the three following alternatives:

\begin{itemize}
    \item \textbf{Local Agreement (LA)} \citep{liu20s_interspeech}: 
    this policy prescribes generating a partial hypothesis from scratch at every newly received audio segment, and emitting it (or only a part of it) if it coincides with one of 
    those generated in the previous time step;
    \item \textbf{Encoder-Decoder Attention (\textsc{EDAtt})} \citep{papi2022attention}: 
    it exploits the cross-attention scores modeling the audio-translation relation to decide whether to emit the words of a partial hypothesis or not. If, for the current word, the sum of the attention scores of the last $\lambda$ received speech frames exceeds a certain threshold $\alpha$ (both $\lambda$ and $\alpha$ are hyperparameters), the emission is 
    delayed because the system needs more context to translate that word. Otherwise, the word is emitted and we proceed to the next word of the hypothesis;
    \item \textbf{\textsc{AlignAtt}} \citep{papi-et-al-2023-alignatt}: as for \textsc{EDAtt}, the cross-attention scores are leveraged to decide what to emit but, in this case, instead of summing the attention scores of the last speech frames, each word is uniquely assigned (or aligned) to the frame having the maximum attention score. If the aligned frame corresponds to one of the last $f$ 
    frames ($f$ being a hyperparameter that controls the latency) the emission is 
    stopped. Otherwise, we proceed to the next word.
\end{itemize}

\subsection{Automatic Subtitling}
\label{subsec:sub}
So far, the adoption of direct ST architectures to address the automatic subtitling task has only been explored in \citep{papi2022direct}.
As a matter of fact, all previous works on the topic \citep{piperidis-etal-2004-multimodal,melero-2006-etitle,matusov-etal-2019-customizing,koponen-etal-2020-mt,bojar-etal-2021-elitr} rely on cascade architectures that usually involve an ASR component to transcribe the input speech, a subtitle segmenter that segments the transcripts into subtitles, a timestamp estimator that predicts the start and times of each subtitle, and an MT model that translates the subtitle transcripts.

Cascaded architectures, however, cannot access information contained in the speech, such as prosody, which related works proved to be an important source of information for the segmentation into subtitles \citep{ktem2019ProsodicPA,Federico2020,9414966,tam22_interspeech}.
The importance of such information has been further verified in \citep{karakanta-etal-2020-42}, which proved that the direct ST models are better in subtitle segmentation compared to the cascade ones.
Another study by \citealt{karakanta-etal-2021-flexibility}, also pointed out the importance of consistency between captions (segmented transcripts) and subtitles (segmented translations), showing that the predicted caption content can also be useful for the translation. Specifically, the authors obtained significant improvements by using a Triangle Transformer-based architecture \citep{anastasopoulos-chiang-2018-tied}
composed of one encoder and two decoders: the first decoder is in charge of emitting the transcripts and the second one is in charge of emitting the translation by also attending to the output embeddings of the predicted transcript.
Therefore, in our submission, based on the findings of the aforementioned work, we inspected the use of both a classic 
single encoder-single decoder architectures, 
as in \citep{papi2022direct}, and of the Triangle architecture for automatic subtitling.

\section{Experimental Setup}

\subsection{Data}

\paragraph{Simultaneous} We developed a pure offline model trained on the same data used for our last year's (constrained) submission \citep{gaido-etal-2022-efficient}. 

\paragraph{Subtitling} We used the same 
data settings of
\citep{papi2022direct}, for which we leverage the multimodal segmenter by \citet{segmenter} to segment into subtitles ST and machine-translated ASR corpora as per \citep{gaido-2020-on-knowledge,Gaido2022DirectST}.\footnote{All the corpora used in \citep{papi2022direct} are allowed ASR and ST training data for the Subtitling task (\url{https://iwslt.org/2023/subtitling\#training-and-data-conditions}). Therefore, our submission has to be considered \enquote{Constrained}.} 
No OpenSubtitles or text-only data were used to train our models.

\subsection{Training Settings}
\label{subsec:train-settings}

All the models used for our participation were implemented using the newly released implementation of the Conformer architecture by \citet{papi2023reproducibility}\footnote{Code available at \url{https://github.com/hlt-mt/FBK-fairseq}} based on Fairseq-ST \citep{wang2020fairseqs2t}. In their paper, the authors analyzed the most popular open-source libraries for speech recognition/translation and found at least one bug affecting all the existing Conformer implementations, therefore claiming the importance of testing code to avoid the propagation of unreliable findings masked by good results. 

\paragraph{Simultaneous} We tested a Conformer-based architecture \citep{gulati20_interspeech} with two configurations: 12 encoder layers and 16 encoder layers. The number of Transformer decoder layers is 6, we set 512 features for the attention layers and 2,048 hidden units for the feed-forward layers. We used 0.1 dropout for the feed-forward layers, attention layers, and convolutional modules. The kernel size was set to 31 for the point- and depth-wise convolutions. 
We trained with the Adam optimizer \citep{adam} by setting $\beta_1=0.9$ and $\beta_2=0.98$, a weight decay of $0.001$, the learning rate to 0.002 using the inverse square-root scheduler with 25,000 warm-up steps. Label smoothed cross-entropy loss (0.1 smoothing factor) was used together with the CTC loss \citep{Graves2006ConnectionistTC} with weight 0.5. We experimented also by applying the CTC compression mechanism \citep{gaido-etal-2021-ctc} to the source input to shrink its dimension and reduce RAM consumption.
Utterance Cepstral Mean and Variance Normalization (CMVN) was applied during training. Also, we leveraged SpecAugment \citep{Park2019} with frequency mask ($F=27$, and $N=2$), time mask ($N=10$, $T=300$, and $p=0.05$), and no time warp.
Both ST training and ASR pre-training were performed with the same settings. The target vocabulary is of size 16,000, and the source vocabulary is of size 10,000, and are both based on SentencePiece \citep{kudo-richardson-2018-sentencepiece}. We differentiate between original and machine-translated training data by pre-pending a tag (\texttt{nomt} and \texttt{mt}, respectively) to the target text as in all our last years' submissions \citep{gaido-etal-2020-end,papi-etal-2021-dealing,gaido-etal-2022-efficient}.
The total batch size was set to 1,280,000 and was performed on 4 NVIDIA A40 GPUs with 40GB of RAM by setting the mini-batch update frequency to 8 and 40,000 maximum tokens. Maximum updates were set to 100,000.

\paragraph{Automatic Subtitling} 
Both the classic encoder-decoder architecture and the triangle architecture are composed of 12 layers of Conformer encoder and 6 layers of Transformer decoder (which is replicated twice in the triangle model). The dimension of the feed-forward layers is 2,048 and $d=512$ in the attention. The kernel size of the point- and depth-wise convolutions in the convolutional modules is 31. The dropout was set to 0.1. CTC loss with compression is added with weight 0.5 to the cross entropy loss with label smoothing (0.1 of smoothing factor) and optimized with Adam ($\beta_1=0.9$, $\beta_2=0.98$). 
The source vocabulary is of size 8,000 and the target vocabulary of size 16,000 (\texttt{<eob>} and \texttt{<eol>} included); 
both are 
obtained by
SentencePiece models.
The ST pre-training was 
done by setting the learning rate to 0.002 with inverse square-root scheduler and 25,000 warm-up updates. The SubST fine-tuning was 
done by setting a constant learning rate of 0.001. A second fine-tuning was 
done with the same 
setting of \citep{papi2022direct}, but we restored
the punctuation of the ASR datasets which do not contain any (i.e., the TEDLIUM corpus \citep{tedlium}) by using \texttt{bert-restore-punctuation},\footnote{\url{https://huggingface.co/felflare/bert-restore-punctuation}} before 
machine-translating and segmenting
the target texts into subtitles.
We trained the standard architecture with 40,000 maximum tokens on 4 NVIDIA A100 GPUs with 40GB of RAM and we set the update frequency to 2. For the triangle architecture, we set maximum tokens to 20,000 to fit the architecture in memory and the update frequency to 4 to hold the same total batch size of 320,000 tokens. Maximum updates were set to 100,000 for both the pre-training and training phases.

\subsection{Evaluation Settings}

\paragraph{Simultaneous} We exploit the SimulEval tool \citep{ma-etal-2020-simuleval}. To be comparable with the previous years, all the results except this year's submission are shown for the SimulEval v1.0.2, which adopts BLEU \citep{post-2018-call}\footnote{case:mixed|eff:no|tok:13a|smooth:exp|version:1.5.1} to measure translation quality and Average Lagging or AL \citep{ma-etal-2019-stacl} to measure latency. Instead, for this year's submission, we adopt the latest version of SimulEval (1.1.0) with BLEU measured with \texttt{sacrebleu} 2.3.0 and we also report Length-Adaptive Average Lagging or LAAL \cite{papi-etal-2022-generation} and Average Token Delay 
or ATD \cite{kano2022average} 
as additional latency metrics.
All the evaluations were run on a single NVIDIA K80 with 12GB of 
RAM, by applying 
global CMVN to audio input, whose features were estimated on the MuST-C v2 training set. Computational aware metrics (\enquote{\_CA}) refer to the single NVIDIA K80 setting and consider also the model computational time in the delay calculation.

\paragraph{Automatic Subtitling} We adopt the following metrics: SubER-cased (henceforth, SubER) \citep{wilken-etal-2022-suber} for overall subtitle quality, Sigma \citep{karakanta2022evaluating} for the subtitle segmentation quality, and BLEU\footnote{case:mixed|eff:no|tok:13a|smooth:exp|version:2.0.0} for translation 
quality. We also compute the conformity percentage of 42 characters per line (CPL) and 21 characters per second (CPS) or reading speed, as 
suggested on the track website.\footnote{\url{https://iwslt.org/2023/subtitling\#automatic-evaluation}} We neglected the conformity computation of the subtitles with more than two lines since our model only produces subtitles with two lines or less, thus being always 100\% conform. Conformity scores are computed by using the script released for the paper \citep{papi2022direct}.\footnote{Script available at: \url{https://github.com/hlt-mt/FBK-fairseq/blob/master/examples/speech_to_text/scripts/subtitle_compliance.py}}
Dev/test audios are segmented with SHAS \citep{tsiamas22_interspeech}. No audio cleaning is applied.

\section{Results}

\subsection{Simultaneous Translation}

Since we directly employ an offline model for the simultaneous inference, we show in Table \ref{tab:offline} the results of the offline ASR pre-training and ST training. Although the model with 12 encoder layers (row 0) 
obtains lower -- hence better -- WER compared to the 16 encoder-layers model (row 1), the highest -- hence better -- BLEU in ST is achieved by the bigger architecture. 
The performance 
is also slightly enhanced by adding the CTC compression (row 3) during training, which is particularly useful also for the SimulST scenario since it speeds up inference (of about 12/15\%).
Therefore, we select this model for the final submission.
Compared to our last year's submission (row 5), our 16 encoder-layers model scores +0.4 BLEU even if, at this time, we have not fine-tuned it on the in-domain (TED talks) datasets. Our model also performs better than the NAIST last year's system (+11.1 BLEU) while is worse (-1.0 BLEU) compared to the last year's SimulST task winner 
CUNI-KIT whose model,
however, 
leveraged large pre-trained models such as wav2vec 2.0 and mBART50. Compared to  
last year's cascade model by UPV,
we score -1.7 
BLEU. This system, however, also outperformed the CUNI-KIT system by 0.7 BLEU points, indicating that a gap between direct and cascade architectures still exists.

\begin{table}[htb]
    \centering
    \small
    \begin{tabular}{r|l|cc}
        \textbf{id} & \textbf{Model} & \textbf{WER\% ($\downarrow$)} & \textbf{BLEU ($\uparrow$)} \\
        \hline
        \hline
        1 & 12 encoder layers & \textbf{9.7} & 31.6 \\
        2 & 16 encoder layers & 9.9 & \textbf{31.9} \\
        3 & \quad + CTC compress. & - & \underline{\textbf{32.1}} \\
        \hline
        4 & CUNI-KIT 2022\textsuperscript{$\dagger$} & - & 33.1 \\
        5 & FBK 2022 & - & 31.7 \\
        6 & NAIST 2022\textsuperscript{$\ddagger$} & - & 21.0 \\
        7 & UPV 2022 (Cascade)* & 9.5 & 33.8 \\
    \end{tabular}
    \caption{Offline results of our Conformer-based architectures on MuST-C v2 tst-COMMON together with the available results of the last year's SimulST competitors. \textsuperscript{$\dagger$}\citep{polak-etal-2022-cuni}, \textsuperscript{$\ddagger$}\citep{fukuda-etal-2022-naist}, *\citep{iranzo-sanchez-etal-2022-mllp}.}
    \label{tab:offline}
\end{table}

In Figure \ref{fig:policies}, we 
show the simultaneous results of the different policies mentioned in Section \ref{subsec:simul} applied to our offline model. The differences in terms of quality-latency trade-off between the LA and both \textsc{EDAtt} and \textsc{AlignAtt} are evident: the last ones outperform the former with an improvement peak of 1.5 BLEU at lower latency (approximately $1s$). Moreover, when the computationally aware AL is considered, \textsc{EDAtt} and \textsc{AlignAtt} are the only policies able to reach a latency $\leq2s$. Regarding the comparison between \textsc{EDAtt} and \textsc{AlignAtt}, \textsc{AlignAtt} can span a latency between 1 and 2.6$s$ ideally (when unlimited computing resources are available), and between 1.8 and 3.7$s$ computationally aware, while \textsc{EDAtt} is limited to a latency of 1.4 to 2.5$s$ ideally, and 2.3 to 3.6$s$ computationally aware.
We hence select 
\textsc{AlignAtt} as it is able to reach
a wider range of latency.

\input{simul/policy_comparison}

Lastly, we compare our policy with the two winning systems of the last two years (2021, and 2022). 
The 2021 winner \citep{liu-etal-2021-ustc} was based on an architecture named Cross Attention Augmented Transducer (CAAT), which was specifically tailored for the SimulST task \citep{liu-etal-2021-cross} and 
still represents the
state of the art in terms of low latency (considering ideal AL only). 
The 2022 winner (CUNI-KIT \citep{polak-etal-2022-cuni})
was based on the wav2vec 2.0 + mBART50 offline architecture reported in Table \ref{tab:offline}, row 4. They applied the LA policy, the same we analyze in Figure \ref{fig:policies}, to the aforementioned architecture for simultaneous inference. The comparison is reported in Figure \ref{fig:main-res}.
As we can see, there is a 1.0-2.0 BLEU difference between our approach and the IWSLT 2022 winner, which is expected since their offline system is superior compared to ours, as already observed in Table \ref{tab:offline}.
Compared to the IWSLT 2021 winner, we observe a performance drop in our system with AL $\leq1.5s$, while 
the situation is opposite
with AL $>1.5s$. 
However, when we look at the computationally aware metrics, the results completely change. Our system 
clearly outperforms the 2021 winner, with a maximum improvement of about 2 BLEU points. Moreover, our system is the only one able to reach a computational aware latency of 2$s$ while, instead, the IWSLT 2022 winner curve starts only at around 3$s$. Therefore, our system is significantly faster and, at around 3$s$, we achieve a relative improvement of more than 3.5 BLEU compared to the IWSLT 2022 winner. 

To sum up, when the computationally aware metric is considered, our approach outperforms 
the winners of both the 2021 and 2022 rounds of the SimulST task.
In addition, in this year's round, the systems are evaluated with the threshold AL $=2s$ and with the new version of SimulEval.\footnote{\url{https://iwslt.org/2023/simultaneous\#ranking}} 
With respect to these settings, our submitted system scores $30.7$ BLEU with AL $=1.89s$ (LAAL $=2.07s$, ATD $=1.80s$).

\input{simul/last_years}

\subsection{Automatic Subtitling}

In Table \ref{tab:single-vs-triangle}, we show a comparison between the standard encoder-decoder and the Triangle architectures for automatic subtitling. The results are computed  on MuST-Cinema \citep{karakanta-etal-2020-must}, the only existing corpus for SubST.
Unfortunately, 
in contrast with the results achieved by \citep{karakanta-etal-2021-flexibility},
we found that the standard architectures perform better on all the considered metrics. While the differences in terms of translation quality are not so big (0.8-9 BLEU drop in both languages), there is a huge gap in the quality of the segmentation into 
subtitles, with the standard model improving by 3.3 and 4.7 Sigma the scores obtained by the Triangle respectively on en-de and en-es.
This is also reflected by a worse SubER score (the lower, the better) of the Triangle, exhibiting a performance drop of, respectively, 0.9 and 1.6 SubER for en-de and en-es compared to the standard architecture. Therefore, we can conclude that the generated 
captions seem not to help with subtitle generation. Rather, they negatively affect subtitle generation to the detriment of segmentation quality. For this reason, we decided to employ the standard encoder-decoder architecture for our participation in the automatic subtitling task.

\begin{table}[htb]
    \small
    \setlength{\tabcolsep}{3pt}
    \centering
    \begin{tabular}{l|ccccc}
        \hline
        \hline
        \multicolumn{6}{c}{\textbf{en-de}} \\
        \hline
        \textbf{Model} & \textbf{SubER} & \textbf{BLEU} & \textbf{Sigma} & \textbf{CPL} & \textbf{CPS} \\
        \hline
        \citep{papi2022direct} & \textbf{59.9} & \textbf{23.4} & \textbf{77.9} & \textbf{86.9} & \textbf{68.9} \\
        Triangle & 60.8 & 22.6 & 74.6  & 84.5 & 67.7 \\
        \hline
        \hline
        \multicolumn{6}{c}{\textbf{en-es}} \\
        \hline
        \textbf{Model} & \textbf{SubER} & \textbf{BLEU} & \textbf{Sigma} & \textbf{CPL} & \textbf{CPS} \\
        \hline
        \citep{papi2022direct} & \textbf{46.8} & \textbf{37.4} & \textbf{81.6} & \textbf{93.2} & \textbf{74.6} \\
        Triangle & 48.4 & 36.5 & 76.9 & 90.3 & 71.7 \\
        \hline
    \end{tabular}
    \caption{Results of the direct ST models standard and Triangle architectures described in Section \ref{subsec:sub} on MuST-Cinema test set for en$\rightarrow$\{de, es\}.}
    \label{tab:single-vs-triangle}
\end{table}

In the following, we present the results of our model on the four dev sets released for the task,\footnote{\url{https://iwslt.org/2023/subtitling\#development-and-evaluation-data}} namely: \textbf{MuST-Cinema or TED} containing TED talks videos, \textbf{EuroParlTV or EPTV} containing recordings related to the European Parliament activities, \textbf{Peloton} containing online fitness classes, and \textbf{ITV Studios or ITV} containing videos from a broad range of programming (drama, entertainment, factual).
For both language pairs (en-de and en-es), Table \ref{tab:sub-dev} shows the results computed with SubER, which is the primary metric used for the task.\footnote{\url{https://iwslt.org/2023/subtitling\#automatic-evaluation}}
As we can see, the 
models fine-tuned on data with restored punctuation score the best results in both languages.
Across the four dev sets, there is a 3.7 SubER improvement for en-de, and 1.7 for en-es.
Moreover, coherently among languages, the TED talks scenario results in the easiest one for our model, as it is in-domain (e.g., MuST-Cinema, based on TED talks, was used to train the model). Conversely, the ITV scenario is the most difficult one since it contains TV series, which is a completely unseen domain for our 
model. Indeed, its data contain a larger amount of background music/noise, as well as dialogues with multiple speakers which are not present in our training data.
In light of the results obtained by the fine-tuned models, we select them for our submission to the automatic subtitling task.

\begin{table}[htb]
    \small
    \setlength{\tabcolsep}{3.8pt}
    \centering
    \begin{tabular}{l|cccc|c}
        \hline
        \hline
        \multicolumn{6}{c}{\textbf{en-de}} \\
        \hline
        \textbf{Model} & \textbf{TED} & \textbf{EPTV} & \textbf{Peloton} & \textbf{ITV} & \textbf{Avg} \\
        \hline
        \citep{papi2022direct} & 72.7 & 82.3 & 84.7 & 88.0 & 81.9 \\
        \quad+ fine-tuning & \textbf{69.4} & \textbf{80.6} & \textbf{79.1} & \textbf{83.7} & \textbf{78.2} \\
        \hline
        \hline
        \multicolumn{5}{c}{\textbf{en-es}} \\
        \hline
        \textbf{Model} & \textbf{TED} & \textbf{EPTV} & \textbf{Peloton} & \textbf{ITV} & \textbf{Avg} \\
        \hline
        \citep{papi2022direct} & 54.8 & 75.3 & 82.3 & 84.1 & 74.1 \\
        \quad+ fine-tuning & \textbf{52.5} & \textbf{73.7} & \textbf{80.3} & \textbf{82.2} & \textbf{72.4} \\
        \hline
    \end{tabular}
    \caption{SubER ($\downarrow$) scores for en$\rightarrow$\{de, es\} of the direct ST models on the four dev sets of the competition. \enquote{fine-tuning} represents the second fine-tuning on data with restored punctuation mentioned in Section \ref{subsec:train-settings}.}
    \label{tab:sub-dev}
\end{table}

\section{Conclusions}

We presented the FBK's systems built to participate in the IWSLT 2023 Evaluation Campaigns for simultaneous speech translation (en-de) and automatic subtitling (en-\{de, es\}). 
Our submissions are characterized by the use of direct speech translation models to address both tasks, without any further modification nor adaptation for the simultaneous task, and with a fine-tuning on subtitle-like translations for the automatic subtitling task. 
Our SimulST system achieves a lower computational-aware latency with up to 3.5 BLEU gain compared to the last two years' winners. Our automatic subtitling system achieves 3.7 and 1.7 SubER improvement on en-de and en-es respectively, compared to the only solution published in the literature based on a direct system.

\section*{Acknowledgements}
This work has been supported by the project \enquote{AI@TN} funded by the Autonomous Province of Trento, Italy.

\bibliography{custom}
\bibliographystyle{acl_natbib}




\end{document}

%% file: simul/policy_comparison.tex
\pgfplotstableread[row sep=\\]{
BLEU	AL AL_CA \\
22.7	1.097  2.458 \\
27.3	1.547  2.634 \\
29.3	1.910  2.894 \\
30.5	2.227  3.150 \\
31.2    2.783   3.689 \\
}\LA

\pgfplotstableread[row sep=\\]{
BLEU	AL AL_CA\\
27.6	1.384  2.263 \\
28.5    1.541   2.483 \\
29.5    1.822   2.761 \\
30.4    2.118   3.139 \\
31.0    2.471   3.589 \\
}\EdAtt

\pgfplotstableread[row sep=\\]{
BLEU	AL AL_CA \\
21.1	1.022  1.840 \\
25.8	1.194  2.055 \\
27.1	1.336  2.198 \\
28.9	1.635  2.537 \\
29.7    1.891   2.826 \\
30.4    2.141   3.124 \\
31.0    2.593   3.704 \\
}\AlignAtt

\pgfplotstableread[row sep=\\]{
BLEU	AL \\
32.1	0 \\
32.1	6.0 \\
}\offline

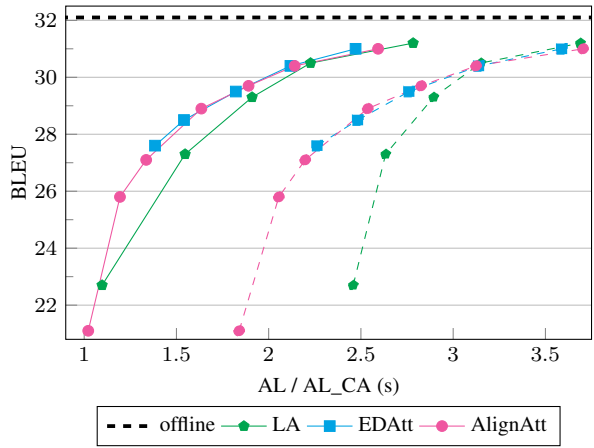
\begin{figure}[t]
\centering
\small
\begin{tikzpicture}
    \begin{axis}[
            ymajorgrids=true,
            xtick pos=left,
            ytick pos=left,
            minor y tick num=1,
            ytick={22,24,26,28,30,32},
            ymax=32.5,
            ymin=20.8,
            xmax=3.75,
            xmin=0.9,
            ylabel=BLEU, xlabel=AL / AL\_CA (s),
            ylabel shift={-3pt},
            width=8.5cm,
            height=6cm,
            xtick=data,
            compat=newest,
            xtick={1,1.5,2,2.5,3,3.5,4},
            legend style={at={(0.5,-0.2)},    
                    anchor=north,legend columns=4},   
        ]
        \addplot[dashed, color=black, line width=0.5mm] table[x=AL,y=BLEU]{\offline};
        \addplot[color=Green, mark=pentagon*] table[x=AL,y=BLEU]{\LA};
        \addplot[dashed, color=Green, mark=pentagon*] table[x=AL_CA,y=BLEU]{\LA};
        \addplot[color=Cerulean, mark=square*] table[x=AL,y=BLEU]{\EdAtt};
        \addplot[dashed, color=Cerulean, mark=square*] table[x=AL_CA,y=BLEU]{\EdAtt};
        \addplot[color=VioletRed, mark=*] table[x=AL,y=BLEU]{\AlignAtt};
        \addplot[dashed, color=VioletRed, mark=*] table[x=AL_CA,y=BLEU]{\AlignAtt};
        \legend{offline,LA,,EDAtt,,AlignAtt}
    \end{axis}
\end{tikzpicture}
\caption{Comparison between the LA, \textsc{EDAtt}, and \textsc{AlignAtt} policies described in Section \ref{subsec:simul} on MuST-C v2 en$\rightarrow$de tst-COMMON. Solid curves represent AL, dashed curves represent AL\_CA.}
\label{fig:policies}
\end{figure}

%% file: simul/last_years.tex
\pgfplotstableread[row sep=\\]{
BLEU	AL AL_CA \\
27.4	0.920   2.330 \\
29.7	1.860  3.660 \\
30.8	2.740  5.050 \\
}\iwsltA

\pgfplotstableread[row sep=\\]{
BLEU	AL AL_CA \\
26.8	0.960  2.940 \\
31.5	1.930  3.710 \\
32.9	3.660  5.540 \\
}\iwsltB

\pgfplotstableread[row sep=\\]{
BLEU	AL AL_CA \\
25.8	1.194  2.055 \\
27.1	1.336  2.198 \\
28.9	1.635  2.537 \\
29.7    1.891   2.826 \\
30.4    2.141   3.124 \\
31.0    2.593   3.704 \\
}\AlignAtt

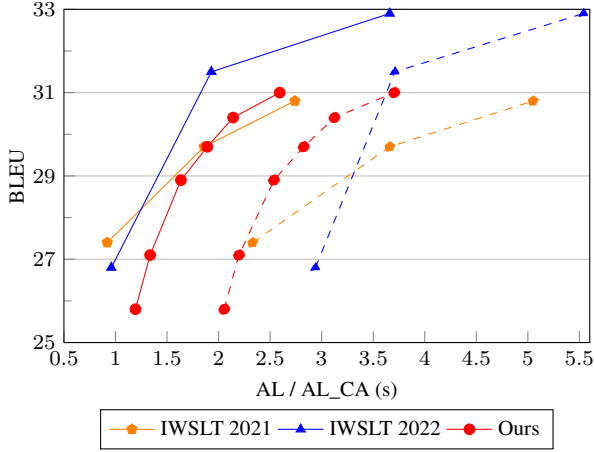
\begin{figure}[t]
\centering
\small
\begin{tikzpicture}
    \begin{axis}[
            ymajorgrids=true,
            xtick pos=left,
            ytick pos=left,
            minor y tick num=1,
            ytick={25,27,29,31,33},
            ymax=33,
            ymin=25,
            xmax=5.6,
            xmin=0.5,
            ylabel=BLEU, xlabel=AL / AL\_CA (s),
            ylabel shift={-3pt},
            width=8.5cm,
            height=6cm,
            xtick=data,
            compat=newest,
            xtick={0.5,1,1.5,2,2.5,3,3.5,4,4.5,5,5.5,6},
            legend style={at={(0.5,-0.2)},    
                    anchor=north,legend columns=4},   
        ]
        \addplot[color=orange, mark=pentagon*] table[x=AL,y=BLEU]{\iwsltA};
        \addplot[dashed, color=orange, mark=pentagon*] table[x=AL_CA,y=BLEU]{\iwsltA};
        \addplot[color=blue, mark=triangle*] table[x=AL,y=BLEU]{\iwsltB};
        \addplot[dashed, color=blue, mark=triangle*] table[x=AL_CA,y=BLEU]{\iwsltB};
        \addplot[color=red, mark=*] table[x=AL,y=BLEU]{\AlignAtt};
        \addplot[dashed, color=red, mark=*] table[x=AL_CA,y=BLEU]{\AlignAtt};
        \legend{IWSLT 2021,,IWSLT 2022,,Ours}
    \end{axis}
\end{tikzpicture}
\caption{Comparison with the 2021 and 2022 winners of the SimulST Evaluation Campaigns MuST-C v2 en$\rightarrow$\{de, es\} tst-COMMON. Solid curves represent AL, dashed curves represent AL\_CA.}
\label{fig:main-res}
\end{figure}